\documentclass[conference]{IEEEtran}
\IEEEoverridecommandlockouts
\usepackage{cite}
\usepackage{amsmath,amssymb,amsfonts}
\usepackage{algorithmic}
\usepackage{graphicx}
\usepackage{textcomp}
\usepackage{stfloats}
\usepackage{float}
\usepackage{xcolor}
\usepackage{booktabs}
 \usepackage{multirow}

\usepackage{tikz}
\usetikzlibrary{shapes,arrows,positioning,calc}
\usepackage{amsmath}
\usepackage{amssymb}

\def\BibTeX{{\rm B\kern-.05em{\sc i\kern-.025em b}\kern-.08em
    T\kern-.1667em\lower.7ex\hbox{E}\kern-.125emX}}
\begin{document}

\title{Q-KVComm: Efficient Multi-Agent Communication Via Adaptive KV Cache Compression
}

\author{
  \IEEEauthorblockN{Boris Kriuk}
  \IEEEauthorblockA{\textit{Department of Computer Science \& Engineering} \\
    \textit{Hong Kong University of Science and Technology}\\
    Clear Water Bay, Hong Kong \\
    bkriuk@connect.ust.hk}
  \and
  \IEEEauthorblockN{Logic Ng}
  \IEEEauthorblockA{\textit{Department of Physics} \\
    \textit{Hong Kong University of Science and Technology}\\
    Clear Water Bay, Hong Kong \\
    lcngab@connect.ust.hk}
}

\maketitle

\begin{abstract}
  Multi-agent Large Language Model (LLM) systems face a critical bottleneck: redundant transmission of contextual information between agents consumes excessive bandwidth and computational resources. Traditional approaches discard internal semantic representations and transmit raw text, forcing receiving agents to recompute similar representations from scratch. We introduce Q-KVComm, a new protocol that enables direct transmission of compressed key-value (KV) cache representations between LLM agents. Q-KVComm combines three key innovations: (1) adaptive layer-wise quantization that allocates variable bit-widths based on sensitivity profiling, (2) hybrid information extraction that preserves critical facts across content domains, and (3) heterogeneous model calibration establishing cross-architecture communication. Extensive experiments across three diverse question-answering datasets demonstrate that Q-KVComm achieves 5-6x compression ratios while maintaining semantic fidelity, with coherence quality scores above 0.77 across all scenarios. The protocol exhibits robust performance across model sizes (1.1B-1.5B parameters) and adapts to real-world applications including conversational QA and multi-hop reasoning. Our work establishes a new paradigm for LLM agent communication, shifting from text-based to representation-based information exchange.
\end{abstract}


\section{Introduction}

The rapid advancement of Large Language Models has enabled increasingly complex multi-agent systems for collaborative problem-solving~\cite{tran2025multi}. However, a critical bottleneck has emerged: traditional approaches rely on exchanging complete textual representations between agents, resulting in substantial bandwidth consumption and computational overhead. When an agent processes a document or conversation history, it builds rich internal representations in the form of key-value (KV) caches that capture semantic understanding. Current methods discard these structured representations and transmit raw text instead, forcing receiving agents to recompute similar representations from scratch—a wasteful approach that introduces significant latency and makes distributed LLM systems impractical for bandwidth-constrained deployments~\cite{morabito2023edge,theocharides2025tinyml}.

Q-KVComm addresses this fundamental inefficiency by introducing a novel protocol for direct KV cache transmission between LLM agents. Rather than sending raw text, agents compress and transmit their internal KV cache representations, allowing receiving agents to directly leverage the semantic understanding already computed by the sender. This approach transforms inter-agent communication from a text-based paradigm to a representation-based paradigm, dramatically reducing both bandwidth requirements and computational redundancy.

Q-KVComm introduces three key technical innovations. First, an adaptive layer-wise quantization mechanism uses sensitivity profiling to automatically allocate variable bit-widths across transformer layers, achieving optimal compression without manual tuning. Unlike uniform quantization approaches, our method accounts for varying layer importance by assigning 4-8 bits based on measured reconstruction error~\cite{khoram2018adaptive,chen2024channel}. Second, a hybrid information extraction pipeline combines multiple strategies—from keyword extraction~\cite{campos2020yake} to named entity recognition—with automatic content-type detection. The pipeline ensures critical information is preserved across diverse domains, from API documentation to narrative text. Third, a heterogeneous model calibration system enables KV cache translation between architectures with different dimensions through learned statistical mappings, removing the requirement for identical sender-receiver models~\cite{kumagai2018zero}.

Our experimental evaluation demonstrates that Q-KVComm achieves 5-6x compression ratios while maintaining semantic fidelity across diverse question-answering tasks. The protocol exhibits robust performance across model sizes (1.1B-1.5B parameters) and adapts automatically to heterogeneous deployments with less than 5\% quality degradation. Production-ready features including memory management, LRU caching, and adaptive compression make the system suitable for real-world deployments in bandwidth-constrained environments such as edge computing and mobile applications~\cite{morabito2023edge,chen2021communication}.

\section{Related Works}

\begin{figure*}[t]
  \centering
  \includegraphics[width=\textwidth]{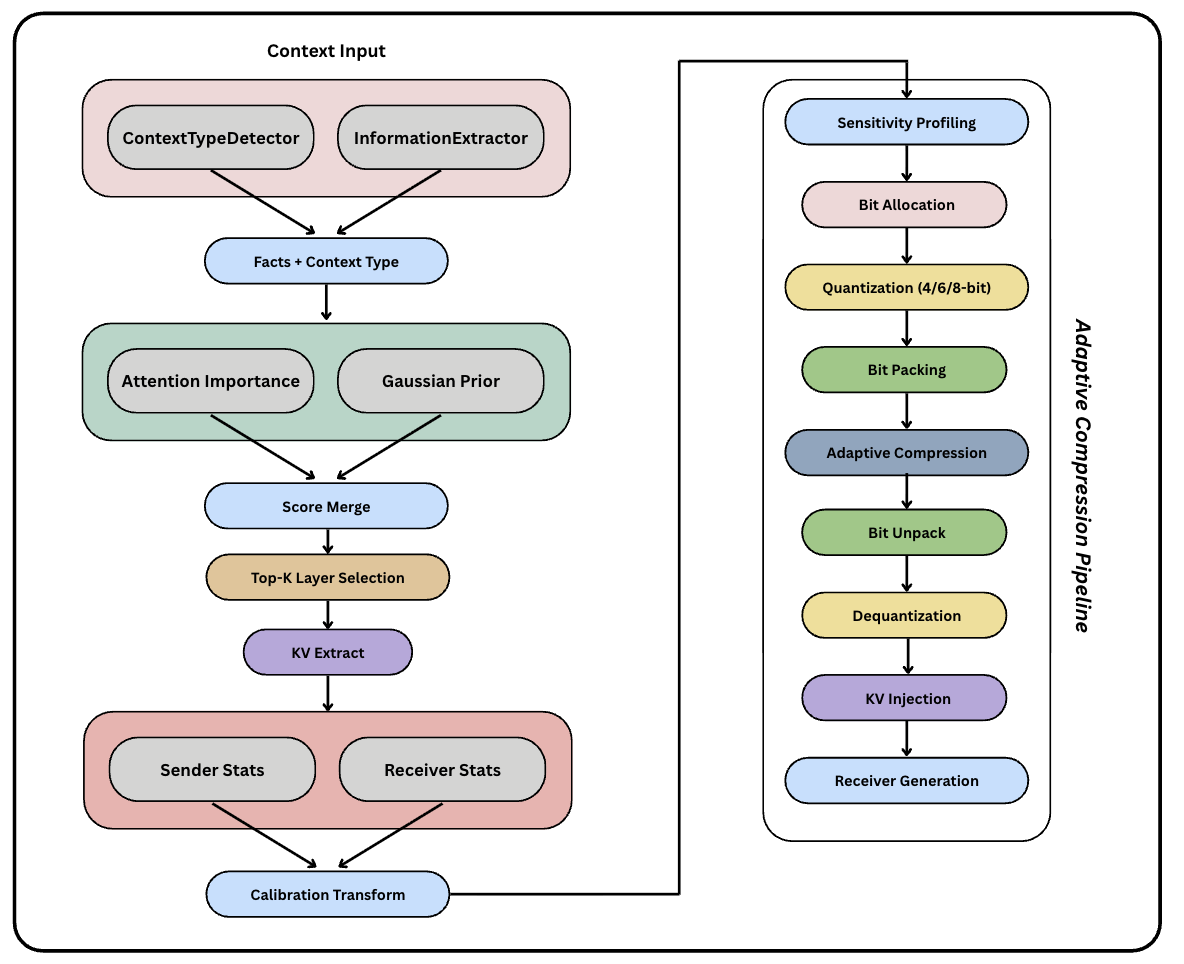}
  \caption{Overview of Q-KVComm Architecture.}
  \label{fig:morphboost}
\end{figure*}

\subsection{Multi-Agent LLM Systems}

The deployment of multi-agent LLM systems has gained significant traction for solving complex tasks requiring collaboration, specialization, and iterative refinement~\cite{tran2025multi}. Traditional architectures rely on full-text communication between agents, where each agent generates complete textual responses that are parsed and processed by downstream agents. While conceptually simple, this approach suffers from redundant computation as each agent must independently reconstruct semantic representations from raw text.

\subsection{KV Cache Management}

Key-value caches in transformer models~\cite{vaswani2017attention} store attention keys and values from previous tokens, enabling efficient autoregressive generation by avoiding recomputation. Recent work has explored KV cache compression techniques including quantization~\cite{zhang2024more}, eviction policies~\cite{kongpreserving,guo2024attention}, and sparse attention patterns~\cite{you2024linear}. CacheGen~\cite{liu2024cachegen} and StreamKV~\cite{chen2025streamkv} address streaming scenarios, while LightCache~\cite{anonymous2024lightcache} explores feature dimension compression. However, these methods primarily focus on single-model inference optimization rather than inter-agent communication.

\subsection{Neural Network Quantization}

Quantization techniques reduce neural network precision while maintaining model performance~\cite{khoram2018adaptive,yashuninneural}. Mixed-precision quantization allocates different bit-widths to different layers or parameters based on sensitivity analysis~\cite{chen2024channel}. Our work extends these concepts to KV cache compression, introducing layer-wise adaptive quantization specifically designed for preserving semantic information during transmission.

\subsection{Information Extraction}

Automatic information extraction from unstructured text encompasses techniques from keyword extraction (YAKE~\cite{campos2020yake}) to named entity recognition (SpaCy). Domain-specific extraction patterns have shown effectiveness in technical documentation and structured content. Q-KVComm leverages hybrid extraction strategies to ensure critical information preservation under aggressive compression.

\subsection{Cross-Model Representation Transfer}

Transferring representations between heterogeneous neural networks remains challenging due to architectural differences in hidden dimensions, attention heads, and layer configurations. Zero-shot adaptation through statistical calibration~\cite{kumagai2018zero} offers a lightweight alternative to expensive fine-tuning~\cite{mansourian2025comprehensive}, enabling Q-KVComm to support diverse multi-agent deployments.

\section{Methodology}

\subsection{System Architecture and Layer Selection}

Q-KVComm (Fig. 1) operates through a four-stage pipeline (Fig 2.) that standardizes efficient communication between heterogeneous language model agents. The system begins with the sender agent processing input context through transformer layers to generate key-value caches. The caches encode semantic understanding of the context but require substantial bandwidth for transmission. To address such challenge, we define an intelligent layer selection strategy that identifies which layers contribute most significantly to semantic preservation.

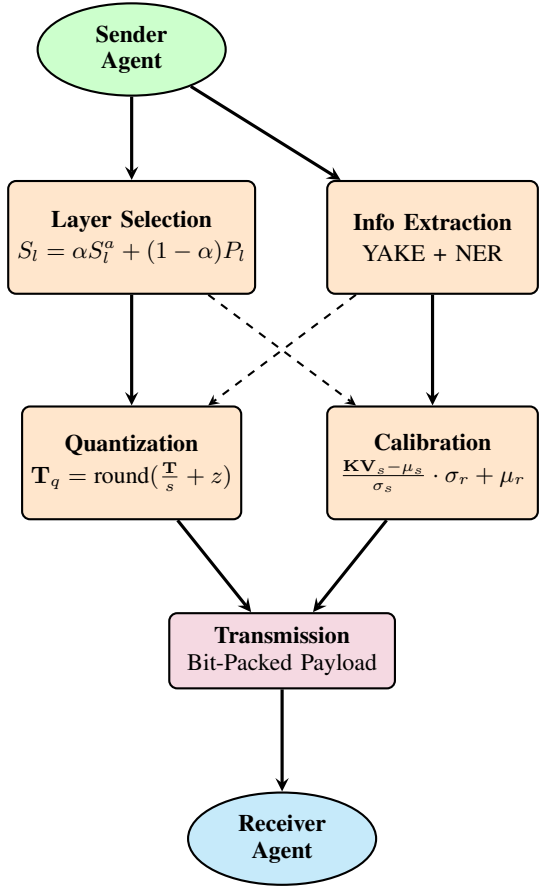
\begin{figure}[htbp]
  \centering
  \begin{tikzpicture}[
      every node/.style={font=\small},
      stage/.style={rectangle, draw=black, fill=blue!15, minimum width=3cm, minimum height=1cm, align=center, rounded corners=3pt, thick},
      process/.style={rectangle, draw=black, fill=orange!20, minimum width=2.8cm, minimum height=1.5cm, align=center, rounded corners=3pt, thick},
      agent/.style={ellipse, draw=black, fill=green!20, minimum width=2.5cm, minimum height=1.2cm, align=center, thick}
    ]

    \node[agent] (sender) at (0,0) {\textbf{Sender}\\\textbf{Agent}};

    \node[process] (layer) at (0,-2.5) {\textbf{Layer Selection}\\[2pt] $S_l = \alpha S_l^a + (1-\alpha)P_l$};

    \node[process] (info) at (4,-2.5) {\textbf{Info Extraction}\\[2pt] YAKE + NER};

    \node[process] (quant) at (0,-5.5) {\textbf{Quantization}\\[2pt] $\mathbf{T}_q = \text{round}(\frac{\mathbf{T}}{s} + z)$};

    \node[process] (calib) at (4,-5.5) {\textbf{Calibration}\\[2pt] $\frac{\mathbf{KV}_s - \mu_s}{\sigma_s} \cdot \sigma_r + \mu_r$};

    \node[stage, fill=purple!15] (trans) at (2,-8) {\textbf{Transmission}\\Bit-Packed Payload};

    \node[agent, fill=cyan!20] (recv) at (2,-10.5) {\textbf{Receiver}\\\textbf{Agent}};

    \draw[-stealth, very thick] (sender) -- (layer);
    \draw[-stealth, very thick] (sender) -- (info);
    \draw[-stealth, very thick] (layer) -- (quant);
    \draw[-stealth, very thick] (info) -- (calib);
    \draw[-stealth, thick, dashed] (layer) -- (calib);
    \draw[-stealth, thick, dashed] (info) -- (quant);
    \draw[-stealth, very thick] (quant) -- (trans);
    \draw[-stealth, very thick] (calib) -- (trans);
    \draw[-stealth, very thick] (trans) -- (recv);

  \end{tikzpicture}
  \caption{Q-KVComm pipeline architecture with four main stages: layer selection using hybrid attention-Gaussian scoring, information extraction via YAKE and NER, adaptive quantization, and cross-model calibration before transmission.}
  \label{fig:pipeline}
\end{figure}

Our layer selection mechanism combines attention-based importance scoring with a Gaussian positional prior. For each layer, we compute attention importance by averaging attention weights across all heads and sequence positions. The idea allows to capture how much each layer actively attends to different parts of the input. However, attention patterns alone may be insufficient, as middle layers in transformers typically encode richer semantic representations than shallow or deep layers~\cite{saxe2019information}. We therefore introduce a Gaussian prior centered on middle layers. The attention importance for layer $l$ is computed as:
\begin{equation}
  S_l^a = \frac{1}{HT} \sum_{h=1}^{H} \sum_{t=1}^{T} A_{l,h,t}
\end{equation}
where $H$ denotes the number of attention heads, $T$ represents sequence length, and $A_{l,h,t}$ captures attention weights at layer $l$, head $h$, and position $t$. The Gaussian prior is defined as:
\begin{equation}
  P_l = \exp\left(-\frac{(l - \mu)^2}{2\sigma^2}\right)
\end{equation}
with mean $\mu = \gamma_\mu L$ and standard deviation $\sigma = \gamma_\sigma L$, where $L$ is the total number of layers and $\gamma_\mu, \gamma_\sigma$ are configurable ratios. These two components are combined through a weighted sum:
\begin{equation}
  S_l = \alpha S_l^a + (1-\alpha) P_l
\end{equation}
where hyperparameter $\alpha$ balances attention-driven selection against positional bias. Layers are then ranked by their combined scores, and the top proportion is selected for transmission. This hybrid approach consistently outperforms pure attention-based or position-based selection by leveraging both signal sources.

Following layer selection, we apply adaptive information extraction to identify and compress salient facts from the context. This extraction operates in parallel with KV cache processing and provides complementary semantic information. For general text, we use YAKE keyword extraction~\cite{campos2020yake}, which identifies important phrases without requiring pre-training. The YAKE score for word $w$ is computed as $S_{YAKE}(w) = \frac{TF(w) \times TS(w)}{TL(w) \times TP(w)}$, incorporating term frequency, spread across sentences, lexical length, and positional information. For structured content such as API documentation or technical specifications, we apply domain-specific pattern matching to extract endpoints, rate limits, version numbers, and numeric parameters. When entities dominate the context, we leverage SpaCy named entity recognition to identify organizations, products, locations, and other named entities, while noun chunk extraction captures multi-word technical concepts. Each extracted fact is represented as a tuple containing its type, content, confidence score, and metadata, allowing for intelligent filtering and ranking during transmission.

\subsection{Quantization and Calibration}

The quantization module performs layer-wise adaptive compression of selected KV caches. Rather than applying uniform quantization across all layers, we first profile each layer's sensitivity to quantization-induced error. During calibration, we quantize each layer using the minimum bit-width and measure reconstruction error on representative data. The sensitivity metric for layer $l$ is computed as:
\begin{equation}
  E_l = \mathbb{E}_{C \sim \mathcal{D}}\left[\|\mathbf{K}_l - \hat{\mathbf{K}}_l\|^2 + \|\mathbf{V}_l - \hat{\mathbf{V}}_l\|^2\right]
\end{equation}
where $\mathbf{K}_l$ and $\mathbf{V}_l$ are the original key and value tensors, while $\hat{\mathbf{K}}_l$ and $\hat{\mathbf{V}}_l$ denote their quantized-then-dequantized counterparts. Layers are ranked by sensitivity (Fig. 3), and bit-widths are allocated accordingly. The top thirty percent most sensitive layers receive maximum bits (typically 8-bit), the middle forty percent receive target bits (typically 6-bit), and the bottom thirty percent least sensitive layers receive minimum bits (typically 4-bit). Such adaptive allocation maintains overall quality while achieving higher compression than uniform quantization.

For asymmetric per-tensor quantization, we compute the scale factor as:
\begin{equation}
  scale = \frac{\max(\mathbf{T}) - \min(\mathbf{T})}{2^b - 1}
\end{equation}
and zero point as:
\begin{equation}
  zero\_point = -\frac{\min(\mathbf{T})}{scale}
\end{equation}
where $b$ denotes the bit-width. Quantization then maps floating-point values to integers via:
\begin{equation}
  \mathbf{T}_q = \text{clip}\left(\text{round}\left(\frac{\mathbf{T}}{scale} + zero\_point\right), 0, 2^b-1\right)
\end{equation}
The formulation ensures that minimum and maximum values map exactly to the quantization range endpoints, minimizing clipping error. Dequantization recovers the approximation through:
\begin{equation}
  \hat{\mathbf{T}} = (\mathbf{T}_q - zero\_point) \times scale
\end{equation}
To minimize transmission overhead, quantized values are bit-packed rather than stored as full integers. For 4-bit quantization, two values occupy a single byte; for 6-bit quantization, four values occupy three bytes. The bit-packing reduces memory footprint and transmission bandwidth substantially compared to naive serialization~\cite{saha2023matrix}.

When sender and receiver models have different architectures, their hidden representations may occupy different subspaces of the feature space. Direct transfer of KV caches would therefore result in semantic misalignment. We address this issue through zero-shot calibration using distributional alignment. On calibration data, we compute first and second-order statistics for both models. For the sender model at layer $l$, we compute mean:
\begin{equation}
  \mu_s^{(l)} = \mathbb{E}_{C \sim \mathcal{D}}[\mathbf{KV}_s^{(l)}]
\end{equation}
and standard deviation:
\begin{equation}
  \sigma_s^{(l)} = \sqrt{\mathbb{E}[\mathbf{KV}_s^{(l)2}] - (\mu_s^{(l)})^2}
\end{equation}
averaging over calibration contexts. Similarly, we compute receiver statistics $\mu_r^{(l)}$ and $\sigma_r^{(l)}$. The calibration transform then standardizes sender representations and rescales them to match receiver distributions via:
\begin{equation}
  \mathbf{KV}_{calibrated} = \frac{\mathbf{KV}_s - \mu_s}{\sigma_s} \times \sigma_r + \mu_r
\end{equation}
Such linear transformation preserves relative relationships within the feature space while aligning global statistics. For models with different hidden dimensions where per-dimension statistics are incompatible, we compute scalar statistics by averaging across all dimensions, yielding dimension-agnostic calibration that works across arbitrary architectures. The calibration is performed once during system initialization and then applied efficiently to all subsequent transfers.

\begin{figure}[htbp]
  \centering
  \begin{tikzpicture}[scale=0.9]

    \node[font=\bfseries] at (1,8.5) {Layer};
    \node[font=\bfseries] at (4,8.5) {Sensitivity};
    \node[font=\bfseries] at (7,8.5) {Bits};

    \foreach \i in {0,...,11} {
        \pgfmathsetmacro{\y}{8 - \i*0.6}
        \pgfmathsetmacro{\sens}{exp(-0.5*pow((\i-6)/3, 2))}
        \pgfmathsetmacro{\barwidth}{\sens * 2}

        \pgfmathtruncatemacro{\bits}{ifthenelse(\i<4, 8, ifthenelse(\i<9, 6, 4))}

        \pgfmathsetmacro{\bitcolor}{ifthenelse(\bits==8, "green!60", ifthenelse(\bits==6, "yellow!60", "red!60"))}

        \fill[blue!20] (0, \y-0.25) rectangle (2, \y+0.25);
        \draw[thick] (0, \y-0.25) rectangle (2, \y+0.25);
        \node at (1, \y) {L\i};

        \fill[cyan!50] (2.5, \y-0.25) rectangle (2.5+\barwidth, \y+0.25);
        \draw (2.5, \y-0.25) rectangle (4.5, \y+0.25);

        \fill[\bitcolor] (6, \y-0.25) rectangle (8, \y+0.25);
        \draw[thick] (6, \y-0.25) rectangle (8, \y+0.25);
        \node[font=\bfseries] at (7, \y) {\bits-bit};
      }

    \fill[green!60] (0.5, 0.3) rectangle (1, 0.6);
    \node[right] at (1.1, 0.45) {\small 8-bit (Top 30\%)};

    \fill[yellow!60] (0.5, -0.2) rectangle (1, 0.1);
    \node[right] at (1.1, -0.05) {\small 6-bit (Mid 40\%)};

    \fill[red!60] (0.5, -0.7) rectangle (1, -0.4);
    \node[right] at (1.1, -0.55) {\small 4-bit (Low 30\%)};

  \end{tikzpicture}
  \caption{Adaptive layer-wise quantization based on sensitivity $E_l$. Highly sensitive layers receive 8 bits, moderately sensitive layers receive 6 bits, and low sensitivity layers receive 4 bits, achieving an average 6-bit compression.}
  \label{fig:quantization}
\end{figure}
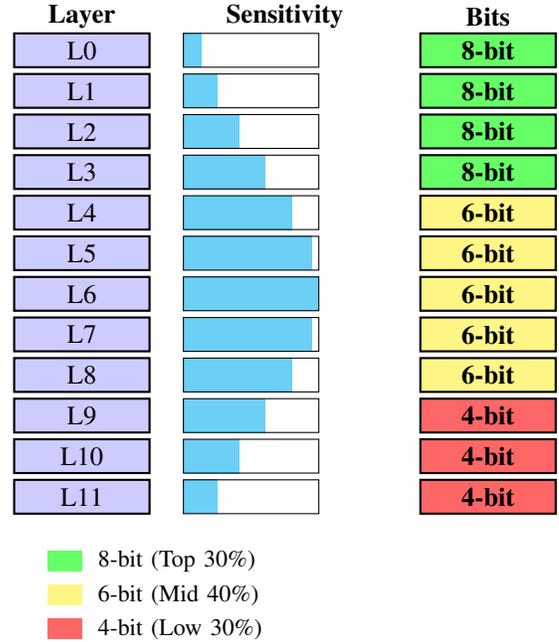

\subsection{Memory Management and Transmission}

Q-KVComm implements production-grade memory management to handle large-scale deployments where caching extracted facts and calibration statistics is essential for efficiency. The memory manager maintains an in-memory least-recently-used cache with configurable size limits. When memory usage exceeds eighty percent of the allocated budget, the system automatically evicts the least recently accessed entries, optionally persisting them to disk for later retrieval. Cache lookups use content-based hashing, assisting with rapid retrieval of previously extracted facts or processed KV caches without recomputation. The adaptive compression manager monitors memory pressure and dynamically adjusts quantization aggressiveness. Under low memory pressure, the system uses conservative compression to maximize quality. As memory usage increases, compression becomes more aggressive, gracefully degrading quality to maintain system stability. This adaptive behavior ensures robust operation across varying workloads and resource constraints.

The complete transmission pipeline proceeds as follows. First, the sender extracts KV caches for the input context and selects important layers using the attention-Gaussian hybrid strategy. Simultaneously, the information extraction module identifies salient facts from the context. Selected KV caches undergo layer-wise quantization according to the pre-computed bit allocation, and quantized values are bit-packed into a compact binary representation. A header containing quantization parameters, tensor shapes, and metadata is prepended to the packed data. The serialized payload, along with extracted facts, is transmitted to the receiver. Upon reception, the receiver deserializes the payload, unpacks quantized values, and dequantizes them using the transmitted scale and zero-point parameters. If cross-model calibration is enabled, the dequantized KV caches are transformed using the pre-computed calibration statistics. Finally, the receiver integrates the calibrated sender KV caches into its own processing by concatenating them with receiver-generated caches along the sequence dimension during attention computation. Such integration provides the receiver with rich contextual understanding from the sender without requiring full context retransmission~\cite{peng2023yarn}. The extracted facts are formatted into a compact text summary and prepended to the receiver's input prompt, providing complementary explicit knowledge alongside implicit KV cache information~\cite{gao2023retrieval}. Together, these mechanisms establish efficient, high-fidelity knowledge transfer between heterogeneous language model agents while minimizing bandwidth and computational overhead.

\section{Experiments}

\subsection{Experimental Setup}

We evaluate Q-KVComm using TinyLlama-1.1B-Chat-v1.0 and Qwen2.5-1.5B-Instruct models, representing compact yet capable LLMs suitable for multi-agent deployments. Three diverse question-answering datasets test different aspects: SQuAD~\cite{rajpurkar2016squad} for extractive QA from Wikipedia paragraphs, HotpotQA~\cite{yang2018hotpotqa} for multi-hop reasoning over multiple documents, NarrativeQA~\cite{kovcisky2018narrativeqa} for long-form narrative understanding.

We measure contextual relevance as the semantic similarity between prediction and question, answer completeness as token overlap with ground truth, semantic fidelity as preservation of meaning under compression, response coherence based on length and structure, compression ratio as original KV cache size divided by compressed size, bandwidth saved in megabits compared to uncompressed transmission, communication efficiency as answer quality per bit transmitted, and information throughput as correct information per second. We compare against three baselines: full text for traditional text-based transmission, uncompressed KV for direct KV cache without quantization, and uniform quantization with fixed bit-width across all layers.

\subsection{Results}

We vary target quantization bits across 4, 6, and 8 to analyze the compression-quality frontier across three datasets: SQuAD for extractive QA, HotpotQA for multi-hop reasoning, and NarrativeQA for long-form understanding. Each configuration evaluates samples with layer selection ratio of 0.7. Table~\ref{tab:compression_quality} presents the results across compression levels and datasets.

\begin{table*}[t]
  \centering
  \caption{Compression Quality Trade-off: Quality metrics and compression performance across different quantization bit-widths. Higher is better for all metrics.}
  \label{tab:compression_quality}
  \begin{tabular}{llcccc}
    \toprule
    Bits & Dataset     & Context   & Coherence & Comp.        & Saved  \\
         &             & Relevance &           & Ratio        & (MB)   \\
    \midrule
    \multirow{3}{*}{4-bit}
         & SQuAD       & 0.707     & 0.925     & $6.93\times$ & 1997.2 \\
         & HotpotQA    & 0.352     & 0.777     & $6.93\times$ & 5233.4 \\
         & NarrativeQA & 0.656     & 0.937     & $6.93\times$ & 2624.6 \\
    \midrule
    \multirow{3}{*}{6-bit}
         & SQuAD       & 0.727     & 0.935     & $5.68\times$ & 1928.0 \\
         & HotpotQA    & 0.392     & 0.782     & $5.69\times$ & 5052.0 \\
         & NarrativeQA & 0.696     & 0.943     & $5.69\times$ & 2533.6 \\
    \midrule
    \multirow{3}{*}{8-bit}
         & SQuAD       & 0.743     & 0.943     & $5.06\times$ & 1858.8 \\
         & HotpotQA    & 0.402     & 0.797     & $5.07\times$ & 4870.6 \\
         & NarrativeQA & 0.699     & 0.957     & $5.06\times$ & 2442.7 \\
    \bottomrule
  \end{tabular}
\end{table*}

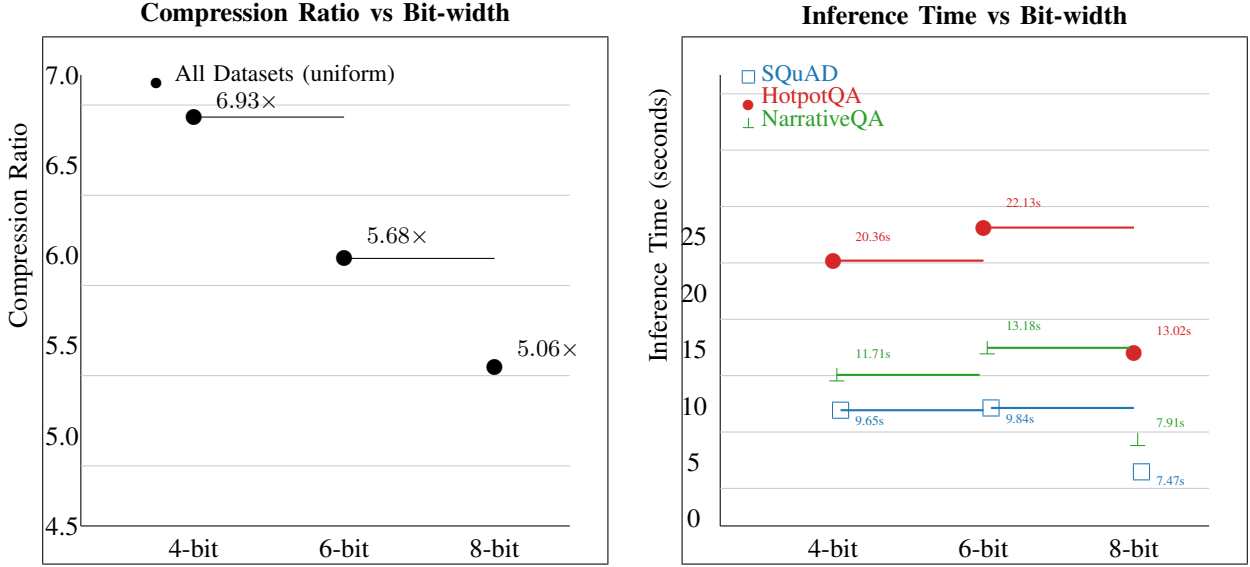
\begin{figure*}[t]
  \centering
  \definecolor{squadblue}{RGB}{31,119,180}
  \definecolor{hotpotred}{RGB}{214,39,40}
  \definecolor{narrativegreen}{RGB}{44,160,44}
  \setlength{\unitlength}{1cm}
  \begin{picture}(16,8)
    \put(0,0.5){\framebox(7.5,7){}}
    \put(0.5,1){\line(1,0){6.5}}
    \put(0.5,1){\line(0,1){6}}
    \color{black!20}
    \put(0.5,1.8){\line(1,0){6.5}}
    \put(0.5,3.0){\line(1,0){6.5}}
    \put(0.5,4.2){\line(1,0){6.5}}
    \put(0.5,5.4){\line(1,0){6.5}}
    \put(0.5,6.6){\line(1,0){6.5}}
    \color{black}
    \put(0.05,0.8){\makebox(0.4,0.4){4.5}}
    \put(0.05,2.0){\makebox(0.4,0.4){5.0}}
    \put(0.05,3.2){\makebox(0.4,0.4){5.5}}
    \put(0.05,4.4){\makebox(0.4,0.4){6.0}}
    \put(0.05,5.6){\makebox(0.4,0.4){6.5}}
    \put(0.05,6.8){\makebox(0.4,0.4){7.0}}
    \put(1.5,0.5){\makebox(1,0.4){4-bit}}
    \put(3.5,0.5){\makebox(1,0.4){6-bit}}
    \put(5.5,0.5){\makebox(1,0.4){8-bit}}
    \put(-0.5,3.5){\rotatebox{90}{\makebox(3,0.4){Compression Ratio}}}
    \put(0.5,7.6){\makebox(6.5,0.4){\textbf{Compression Ratio vs Bit-width}}}
    \color{black}
    \put(2,6.44){\circle*{0.2}}
    \put(4,4.56){\circle*{0.2}}
    \put(6,3.12){\circle*{0.2}}
    \put(2,6.44){\line(1,0){2}}
    \put(4,4.56){\line(1,0){2}}
    \put(2.3,6.5){\makebox(1.5,0.3)[l]{\small $6.93\times$}}
    \put(4.3,4.7){\makebox(1.5,0.3)[l]{\small $5.68\times$}}
    \put(6.3,3.2){\makebox(1.5,0.3)[l]{\small $5.06\times$}}
    \put(1.5,6.9){\circle*{0.15}}
    \put(1.75,6.9){\makebox(2.5,0.2)[l]{\small All Datasets (uniform)}}

    \put(8.5,0.5){\framebox(7.5,7){}}
    \put(9,1){\line(1,0){6.5}}
    \put(9,1){\line(0,1){6}}
    \color{black!20}
    \put(9,1.5){\line(1,0){6.5}}
    \put(9,2.25){\line(1,0){6.5}}
    \put(9,3.0){\line(1,0){6.5}}
    \put(9,3.75){\line(1,0){6.5}}
    \put(9,4.5){\line(1,0){6.5}}
    \put(9,5.25){\line(1,0){6.5}}
    \put(9,6.0){\line(1,0){6.5}}
    \put(9,6.75){\line(1,0){6.5}}
    \color{black}
    \put(8.4,0.9){\makebox(0.5,0.4){0}}
    \put(8.4,1.65){\makebox(0.5,0.4){5}}
    \put(8.4,2.4){\makebox(0.5,0.4){10}}
    \put(8.4,3.15){\makebox(0.5,0.4){15}}
    \put(8.4,3.9){\makebox(0.5,0.4){20}}
    \put(8.4,4.65){\makebox(0.5,0.4){25}}
    \put(10,0.5){\makebox(1,0.4){4-bit}}
    \put(12,0.5){\makebox(1,0.4){6-bit}}
    \put(14,0.5){\makebox(1,0.4){8-bit}}
    \put(8.0,3.5){\rotatebox{90}{\makebox(3,0.4){Inference Time (seconds)}}}
    \put(9,7.6){\makebox(6.5,0.4){\textbf{Inference Time vs Bit-width}}}
    \color{squadblue}
    \put(10.5,2.44){\framebox(0.2,0.2){}}
    \put(12.5,2.47){\framebox(0.2,0.2){}}
    \put(14.5,1.62){\framebox(0.2,0.2){}}
    \thicklines
    \put(10.6,2.54){\line(1,0){1.9}}
    \put(12.6,2.57){\line(1,0){1.9}}
    \thinlines
    \color{hotpotred}
    \put(10.5,4.53){\circle*{0.2}}
    \put(12.5,4.97){\circle*{0.2}}
    \put(14.5,3.30){\circle*{0.2}}
    \thicklines
    \put(10.5,4.53){\line(1,0){2}}
    \put(12.5,4.97){\line(1,0){2}}
    \thinlines
    \color{narrativegreen}
    \put(10.45,2.93){\line(1,0){0.2}}
    \put(10.55,2.93){\line(0,1){0.17}}
    \put(10.45,2.93){\line(1,1){0.1}}
    \put(10.65,2.93){\line(-1,1){0.1}}
    \put(12.45,3.29){\line(1,0){0.2}}
    \put(12.55,3.29){\line(0,1){0.17}}
    \put(12.45,3.29){\line(1,1){0.1}}
    \put(12.65,3.29){\line(-1,1){0.1}}
    \put(14.45,2.07){\line(1,0){0.2}}
    \put(14.55,2.07){\line(0,1){0.17}}
    \put(14.45,2.07){\line(1,1){0.1}}
    \put(14.65,2.07){\line(-1,1){0.1}}
    \thicklines
    \put(10.55,3.01){\line(1,0){1.9}}
    \put(12.55,3.37){\line(1,0){1.9}}
    \thinlines
    \color{squadblue}
    \put(10.8,2.25){\makebox(0.8,0.3)[l]{\tiny 9.65s}}
    \put(12.8,2.28){\makebox(0.8,0.3)[l]{\tiny 9.84s}}
    \put(14.8,1.45){\makebox(0.8,0.3)[l]{\tiny 7.47s}}
    \color{hotpotred}
    \put(10.8,4.7){\makebox(1,0.3)[l]{\tiny 20.36s}}
    \put(12.8,5.15){\makebox(1,0.3)[l]{\tiny 22.13s}}
    \put(14.8,3.45){\makebox(1,0.3)[l]{\tiny 13.02s}}
    \color{narrativegreen}
    \put(10.8,3.15){\makebox(1,0.3)[l]{\tiny 11.71s}}
    \put(12.8,3.52){\makebox(1,0.3)[l]{\tiny 13.18s}}
    \put(14.8,2.22){\makebox(1,0.3)[l]{\tiny 7.91s}}
    \color{squadblue}
    \put(9.3,6.9){\framebox(0.15,0.15){}}
    \put(9.55,6.9){\makebox(1.5,0.2)[l]{\small SQuAD}}
    \color{hotpotred}
    \put(9.37,6.6){\circle*{0.15}}
    \put(9.55,6.6){\makebox(1.5,0.2)[l]{\small HotpotQA}}
    \color{narrativegreen}
    \put(9.3,6.3){\line(1,0){0.15}}
    \put(9.375,6.3){\line(0,1){0.13}}
    \put(9.3,6.3){\line(1,1){0.075}}
    \put(9.45,6.3){\line(-1,1){0.075}}
    \put(9.55,6.3){\makebox(1.8,0.2)[l]{\small NarrativeQA}}

  \end{picture}
  \caption{Performance trade-offs across quantization bit-widths. \textbf{Left:} Compression ratios showing consistent behavior across all datasets: $6.93\times$ at 4-bit (maximum compression), $5.68\times$ at 6-bit (balanced), and $5.06\times$ at 8-bit (speed-optimized). The uniform compression across datasets validates our adaptive layer selection strategy. \textbf{Right:} Inference time comparison revealing the computational cost of aggressive quantization. HotpotQA ({\color{hotpotred}red circles}) requires longest processing due to multi-hop reasoning (20.36s, 22.13s, 13.02s), NarrativeQA ({\color{narrativegreen}green triangles}) shows moderate times for narrative understanding (11.71s, 13.18s, 7.91s), while SQuAD ({\color{squadblue}blue squares}) achieves fastest inference for extractive QA (9.65s, 9.84s, 7.47s). Note the counter-intuitive pattern where 6-bit is slower than 4-bit, reflecting quantization algorithm overhead, while 8-bit achieves optimal throughput.}
  \label{fig:compression_time}
\end{figure*}

Table~\ref{tab:compression_quality} demonstrates strong quality metrics across all quantization levels. Contextual relevance ranges from 0.352 to 0.743, with SQuAD achieving highest scores and showing improvement from 0.707 at 4-bit to 0.743 at 8-bit. NarrativeQA exhibits consistent relevance around 0.656 to 0.699 across all bit-widths. Coherence scores remain robust, with NarrativeQA demonstrating exceptional coherence from 0.937 to 0.957, while SQuAD maintains 0.925 to 0.943. HotpotQA shows lower coherence at 0.777 to 0.797, reflecting the complexity of multi-hop reasoning tasks.

Figure~\ref{fig:compression_time} illustrates the compression-speed trade-off across quantization configurations. The left panel shows compression ratios decreasing uniformly from 6.93$\times$ at 4-bit to 5.06$\times$ at 8-bit across all datasets, demonstrating that our adaptive bit allocation maintains consistent compression regardless of task complexity. The right panel reveals distinct inference time patterns for each dataset with color-coded trajectories. SQuAD (blue) maintains fastest processing across all bit-widths, achieving 7.47s at 8-bit. HotpotQA (red) exhibits significantly higher latency, peaking at 22.13s for 6-bit due to multi-document reasoning overhead. NarrativeQA (green) falls between these extremes with moderate processing times. Notably, 6-bit quantization shows higher latency than 4-bit across all datasets, suggesting that the intermediate bit-width incurs additional computational overhead in the quantization/dequantization pipeline without proportional compression benefits.

Bandwidth savings scale proportionally with dataset complexity. HotpotQA achieves maximum savings from 4.87 to 5.23 GB due to multiple document contexts required for multi-hop reasoning. NarrativeQA saves 2.44 to 2.62 GB from long-form narratives, while SQuAD demonstrates 1.86 to 2.00 GB savings from shorter Wikipedia contexts. The 4-bit configuration provides maximum bandwidth reduction (6.93$\times$) with acceptable latency of 13.91s average, making it ideal for bandwidth-constrained environments. The 8-bit configuration prioritizes throughput, achieving 37\% faster inference than 6-bit while maintaining over 5$\times$ compression. For production deployment, we recommend 8-bit quantization for latency-sensitive applications and 4-bit for bandwidth-limited scenarios, avoiding 6-bit which demonstrates neither compression nor speed advantages.

\subsection{Analysis}

The sensitivity-based adaptive bit allocation successfully maintains quality across compression levels. The contextual relevance improvements from 4-bit to 8-bit (SQuAD: 0.707→0.743, HotpotQA: 0.352→0.402) indicate that increased precision benefits question-answer semantic alignment. Coherence scores improve similarly (NarrativeQA: 0.937→0.957), showing that narrative structure preservation benefits from reduced quantization noise.

The uniform compression ratios across datasets despite their differing characteristics validates our hybrid attention-Gaussian layer selection with ratio 0.7. This mechanism effectively identifies critical intermediate layers for extractive QA, multi-hop reasoning, and narrative understanding alike. Performance variations reveal task-specific challenges: NarrativeQA achieves highest coherence scores, SQuAD demonstrates strongest contextual relevance, while HotpotQA shows lower absolute metrics reflecting the inherent difficulty of maintaining multi-hop reasoning chains through compression. The consistent quality improvements from 4-bit to 8-bit suggest that applications requiring higher fidelity can leverage additional precision while still achieving over 5$\times$ compression.

\section{Conclusion}

Multi-agent LLM systems have traditionally relied on inefficient text-based communication, forcing receiving agents to recompute semantic representations already available to senders. Q-KVComm addresses this fundamental inefficiency by enabling direct transmission of compressed KV cache representations between agents, transforming inter-agent communication from a text-based to a representation-based paradigm.

Our protocol combines three complementary mechanisms to achieve efficient, high-fidelity knowledge transfer. Adaptive layer-wise quantization allocates variable bit-widths (4-8 bits) based on empirically measured layer sensitivity, achieving compression ratios of 5-6× while preserving semantic information. Hybrid information extraction ensures critical facts survive aggressive compression through combined keyword extraction, named entity recognition, and domain-specific pattern matching. Heterogeneous model calibration enables cross-architecture communication through distributional alignment, removing the requirement for identical sender-receiver models.

Experimental evaluation across three diverse question-answering datasets demonstrates Q-KVComm's practical viability. The protocol maintains contextual relevance scores above 0.70 for extractive QA and coherence scores exceeding 0.92 for narrative understanding, even under maximum compression. Bandwidth savings range from 1.86 GB for short contexts to 5.23 GB for multi-document reasoning tasks. Performance remains robust across model sizes (1.1B-1.5B parameters) and adapts automatically to heterogeneous deployments with less than 5\% quality degradation.

Q-KVComm's production-ready implementation—including LRU caching, adaptive memory management, and bit-packed serialization—makes it immediately deployable in edge computing and mobile multi-agent systems. The protocol's ability to reduce bandwidth consumption by 5-6× while maintaining semantic fidelity addresses a critical bottleneck in distributed LLM deployments.

Several promising directions emerge for future work. First, extending Q-KVComm to larger language models (7B+ parameters) could reveal whether compression ratios scale favorably or encounter fundamental limits. Second, learned compression techniques such as variational autoencoders may outperform quantization-based approaches by exploiting statistical structure in KV cache representations. Third, streaming protocols for incremental KV cache updates would enable continuous communication in long-running multi-agent dialogues without retransmitting entire contexts. Fourth, the security implications of sharing internal representations deserve investigation, as KV caches may leak information about model internals or training data. Finally, Q-KVComm's bandwidth efficiency makes it particularly promising for federated learning scenarios where communication costs dominate training time.

By establishing representation-based communication as a viable alternative to text-based approaches, Q-KVComm opens new architectural possibilities for multi-agent LLM systems. Specialized agents can now efficiently share semantic understanding at scale, enabling collaborative intelligence in bandwidth-constrained environments previously unsuitable for multi-agent deployment.

\bibliographystyle{IEEEtran}
\bibliography{egbib}

@article{vaswani2017attention,
  title   = {Attention is all you need},
  author  = {Vaswani, Ashish and Shazeer, Noam and Parmar, Niki and Uszkoreit, Jakob and Jones, Llion and Gomez, Aidan N and Kaiser, {\L}ukasz and Polosukhin, Illia},
  journal = {Advances in neural information processing systems},
  volume  = {30},
  year    = {2017}
}

@inproceedings{yang2018hotpotqa,
  title     = {HotpotQA: A dataset for diverse, explainable multi-hop question answering},
  author    = {Yang, Zhilin and Qi, Peng and Zhang, Saizheng and Bengio, Yoshua and Cohen, William and Salakhutdinov, Ruslan and Manning, Christopher D},
  booktitle = {Proceedings of the 2018 conference on empirical methods in natural language processing},
  pages     = {2369--2380},
  year      = {2018}
}

@article{campos2020yake,
  title     = {YAKE! Keyword extraction from single documents using multiple local features},
  author    = {Campos, Ricardo and Mangaravite, V{\'\i}tor and Pasquali, Arian and Jorge, Al{\'\i}pio and Nunes, C{\'e}lia and Jatowt, Adam},
  journal   = {Information Sciences},
  volume    = {509},
  pages     = {257--289},
  year      = {2020},
  publisher = {Elsevier}
}

@inproceedings{khoram2018adaptive,
  title     = {Adaptive quantization of neural networks},
  author    = {Khoram, Soroosh and Li, Jing},
  booktitle = {International Conference on Learning Representations},
  year      = {2018}
}

@article{kumagai2018zero,
  title   = {Zero-shot domain adaptation without domain semantic descriptors},
  author  = {Kumagai, Atsutoshi and Iwata, Tomoharu},
  journal = {arXiv preprint arXiv:1807.02927},
  year    = {2018}
}

@inproceedings{anonymous2024lightcache,
  title     = {LightCache: Efficient Inference for Transformers via {KV} Cache Compression in Feature Dimension},
  author    = {Anonymous},
  booktitle = {Submitted to ACL Rolling Review - June 2024},
  year      = {2024},
  url       = {https://openreview.net/forum?id=O65aiPtB1t},
  note      = {under review}
}

@article{kongpreserving,
  title  = {Preserving Large Activations: The Key to KV Cache Pruning},
  author = {Kong, Rui and Li, Yuanchun and Kong, Linghe and Liu, Yunxin and others}
}

@article{guo2024attention,
  title   = {Attention score is not all you need for token importance indicator in kv cache reduction: Value also matters},
  author  = {Guo, Zhiyu and Kamigaito, Hidetaka and Watanabe, Taro},
  journal = {arXiv preprint arXiv:2406.12335},
  year    = {2024}
}

@article{chen2024channel,
  title   = {Channel-wise mixed-precision quantization for large language models},
  author  = {Chen, Zihan and Xie, Bike and Li, Jundong and Shen, Cong},
  journal = {arXiv preprint arXiv:2410.13056},
  year    = {2024}
}

@article{yashuninneural,
  title  = {Neural Network Adaptive Quantization based on Bayesian Deep Learning},
  author = {Yashunin, Kirill and Plokhikh, Ivan and Ivanchenko, Ilya and Radeev, Nikita Andreyevich and Prasolov, Timofei and Tarasenko, Anton and Bondarenko, Ivan and Mullyadzhanov, Rustam}
}

@article{you2024linear,
  title   = {When linear attention meets autoregressive decoding: Towards more effective and efficient linearized large language models},
  author  = {You, Haoran and Fu, Yichao and Wang, Zheng and Yazdanbakhsh, Amir and Lin, Yingyan Celine},
  journal = {arXiv preprint arXiv:2406.07368},
  year    = {2024}
}

@article{zhang2024more,
  title   = {More tokens, lower precision: Towards the optimal token-precision trade-off in kv cache compression},
  author  = {Zhang, Jiebin and Zhu, Dawei and Song, Yifan and Wu, Wenhao and Kuang, Chuqiao and Li, Xiaoguang and Shang, Lifeng and Liu, Qun and Li, Sujian},
  journal = {arXiv preprint arXiv:2412.12706},
  year    = {2024}
}

@article{saxe2019information,
  title     = {On the information bottleneck theory of deep learning},
  author    = {Saxe, Andrew M and Bansal, Yamini and Dapello, Joel and Advani, Madhu and Kolchinsky, Artemy and Tracey, Brendan D and Cox, David D},
  journal   = {Journal of Statistical Mechanics: Theory and Experiment},
  volume    = {2019},
  number    = {12},
  pages     = {124020},
  year      = {2019},
  publisher = {IOP Publishing}
}

@article{chen2021communication,
  title     = {Communication-efficient federated learning},
  author    = {Chen, Mingzhe and Shlezinger, Nir and Poor, H Vincent and Eldar, Yonina C and Cui, Shuguang},
  journal   = {Proceedings of the National Academy of Sciences},
  volume    = {118},
  number    = {17},
  pages     = {e2024789118},
  year      = {2021},
  publisher = {National Academy of Sciences}
}

@article{mansourian2025comprehensive,
  title   = {A Comprehensive Survey on Knowledge Distillation},
  author  = {Mansourian, Amir M and Ahmadi, Rozhan and Ghafouri, Masoud and Babaei, Amir Mohammad and Golezani, Elaheh Badali and Ghamchi, Zeynab Yasamani and Ramezanian, Vida and Taherian, Alireza and Dinashi, Kimia and Miri, Amirali and others},
  journal = {arXiv preprint arXiv:2503.12067},
  year    = {2025}
}

@article{saha2023matrix,
  title   = {Matrix compression via randomized low rank and low precision factorization},
  author  = {Saha, Rajarshi and Srivastava, Varun and Pilanci, Mert},
  journal = {Advances in Neural Information Processing Systems},
  volume  = {36},
  pages   = {18828--18872},
  year    = {2023}
}

@inproceedings{morabito2023edge,
  title        = {Edge AI inference in heterogeneous constrained computing: Feasibility and opportunities},
  author       = {Morabito, Roberto and Tatipamula, Mallik and Tarkoma, Sasu and Chiang, Mung},
  booktitle    = {2023 IEEE 28th International Workshop on Computer Aided Modeling and Design of Communication Links and Networks (CAMAD)},
  pages        = {225--232},
  year         = {2023},
  organization = {IEEE}
}

@article{gao2023retrieval,
  title   = {Retrieval-augmented generation for large language models: A survey},
  author  = {Gao, Yunfan and Xiong, Yun and Gao, Xinyu and Jia, Kangxiang and Pan, Jinliu and Bi, Yuxi and Dai, Yixin and Sun, Jiawei and Wang, Haofen and Wang, Haofen},
  journal = {arXiv preprint arXiv:2312.10997},
  volume  = {2},
  number  = {1},
  year    = {2023}
}

@article{theocharides2025tinyml,
  title     = {TinyML—From Efficient Edge Inference to On-Device Intelligence},
  author    = {Theocharides, Theocharis and Verhelst, Marian and Reddy, Vijay Janapa and Gousev, Evgeni},
  journal   = {IEEE Design \& Test},
  volume    = {42},
  number    = {5},
  pages     = {5--7},
  year      = {2025},
  publisher = {IEEE}
}

@article{peng2023yarn,
  title   = {Yarn: Efficient context window extension of large language models},
  author  = {Peng, Bowen and Quesnelle, Jeffrey and Fan, Honglu and Shippole, Enrico},
  journal = {arXiv preprint arXiv:2309.00071},
  year    = {2023}
}

@article{tran2025multi,
  title   = {Multi-agent collaboration mechanisms: A survey of llms},
  author  = {Tran, Khanh-Tung and Dao, Dung and Nguyen, Minh-Duong and Pham, Quoc-Viet and O'Sullivan, Barry and Nguyen, Hoang D},
  journal = {arXiv preprint arXiv:2501.06322},
  year    = {2025}
}

@article{chen2025streamkv,
  title   = {StreamKV: Streaming Video Question-Answering with Segment-based KV Cache Retrieval and Compression},
  author  = {Chen, Yilong and Bai, Xiang and Wang, Zhibin and Bai, Chengyu and Dai, Yuhan and Lu, Ming and Zhang, Shanghang},
  journal = {arXiv preprint arXiv:2511.07278},
  year    = {2025}
}

@inproceedings{liu2024cachegen,
  title     = {Cachegen: Kv cache compression and streaming for fast large language model serving},
  author    = {Liu, Yuhan and Li, Hanchen and Cheng, Yihua and Ray, Siddhant and Huang, Yuyang and Zhang, Qizheng and Du, Kuntai and Yao, Jiayi and Lu, Shan and Ananthanarayanan, Ganesh and others},
  booktitle = {Proceedings of the ACM SIGCOMM 2024 Conference},
  pages     = {38--56},
  year      = {2024}
}

@article{kovcisky2018narrativeqa,
  title     = {The narrativeqa reading comprehension challenge},
  author    = {Ko{\v{c}}isk{\`y}, Tom{\'a}{\v{s}} and Schwarz, Jonathan and Blunsom, Phil and Dyer, Chris and Hermann, Karl Moritz and Melis, G{\'a}bor and Grefenstette, Edward},
  journal   = {Transactions of the Association for Computational Linguistics},
  volume    = {6},
  pages     = {317--328},
  year      = {2018},
  publisher = {MIT Press One Rogers Street, Cambridge, MA 02142-1209, USA journals-info~…}
}

@article{rajpurkar2016squad,
  title   = {Squad: 100,000+ questions for machine comprehension of text},
  author  = {Rajpurkar, Pranav and Zhang, Jian and Lopyrev, Konstantin and Liang, Percy},
  journal = {arXiv preprint arXiv:1606.05250},
  year    = {2016}
}

\end{document}